\documentclass[accepted]{clear2025} 

\usepackage{booktabs} 
\usepackage{siunitx}  
\usepackage{amsmath}
\usepackage{amssymb}
\usepackage{graphicx}
\usepackage{hyperref}
\hypersetup{
    colorlinks=true,
    linkcolor=blue,
    filecolor=magenta,      
    urlcolor=cyan,
    citecolor=blue,
}
\usepackage{times}
\usepackage{multirow}

\usepackage[capitalize,noabbrev]{cleveref}

\newcommand{\indep}{\perp \!\!\! \perp} 

\newtheorem{assumption}{Assumption}

\title[LANCA: Latent Additive Noise Model Causal Autoencoder]{Towards Unsupervised Causal Representation Learning via Latent Additive Noise Model Causal Autoencoders}

\clearauthor{%
 \Name{Hans Jarett J. Ong} \Email{ong.hans\_jarett.ol5@naist.ac.jp}\\
 \addr Nara Institute of Science and Technology, Nara, Japan%
 \AND
 \Name{Brian Godwin S. Lim} \Email{brian.lim@naist.ac.jp}\\
 \addr Nara Institute of Science and Technology, Nara, Japan;\\
       Kyoto University, Kyoto, Japan%
 \AND
 \Name{Dominic Dayta} \Email{dominic.dayta.da4@is.naist.jp}\\
 \addr Nara Institute of Science and Technology, Nara, Japan%
 \AND
 \Name{Renzo Roel P. Tan} \Email{rr.tan@is.naist.jp}\\
 \addr Nara Institute of Science and Technology, Nara, Japan;\\
       Ateneo de Manila University, Metro Manila, Philippines%
 \AND
 \Name{Kazushi Ikeda} \Email{kazushi@is.naist.jp}\\
 \addr Nara Institute of Science and Technology, Nara, Japan%
}

\begin{document}

\maketitle
\thispagestyle{empty}
\begin{abstract}
Unsupervised representation learning seeks to recover latent generative factors, yet standard methods relying on statistical independence often fail to capture causal dependencies. A central challenge is identifiability: as established in disentangled representation learning and nonlinear ICA literature, disentangling causal variables from observational data is impossible without supervision, auxiliary signals, or strong inductive biases. In this work, we propose the \textbf{Latent Additive Noise Model Causal Autoencoder (LANCA)} to operationalize the Additive Noise Model (ANM) as a strong inductive bias for unsupervised discovery. Theoretically, we prove that while the ANM constraint does not guarantee unique identifiability in the general mixing case, it resolves component-wise indeterminacy by restricting the admissible transformations from arbitrary diffeomorphisms to the affine class. 
Methodologically, arguing that the stochastic encoding inherent to VAEs obscures the structural residuals required for latent causal discovery, LANCA employs a \textbf{deterministic Wasserstein Auto-Encoder (WAE)} coupled with a \textbf{differentiable ANM Layer}. This architecture transforms residual independence from a passive assumption into an explicit optimization objective. Empirically, LANCA outperforms state-of-the-art baselines on synthetic physics benchmarks (Pendulum, Flow), and on photorealistic environments (CANDLE), where it demonstrates superior robustness to spurious correlations arising from complex background scenes.
\end{abstract}


\section{Introduction}

Unsupervised representation learning aims to recover the true latent generative factors of high-dimensional data, such as images or text. While standard feature extractors often yield entangled representations, \textbf{disentangled representation learning (DRL)} seeks to isolate distinct semantic factors \citep{Wang2022DisentangledRL}. However, DRL typically relies on the assumption of statistical independence \citep{higgins2017betavae, kim18b-factorvae}. This assumption is frequently violated in physical systems, where factors often share causal dependencies (e.g., a pendulum's angle \textit{causes} the shadow position) \citep{Yang2020CausalVAESC}. Furthermore, \citet{locatello19a-assumptions} demonstrated that without supervision or specific inductive biases, unsupervised disentanglement based on independence is theoretically impossible.

To address these limitations, recent \textbf{causal representation learning (CRL)} \citep{Scholkopf2021TowardsCR} methods like CausalVAE \citep{Yang2020CausalVAESC} and DEAR \citep{Shen2020WeaklySD} have explored replacing the independence assumption with a structural causal model (SCM) prior. However, these methods often rely on supervision to resolve identifiability. To bridge this gap, we introduce the \textbf{Latent Additive Noise Model Causal Autoencoder (LANCA)}. Inspired by causal discovery results which show that Additive Noise Models (ANMs) are identifiable from observational data \citep{hoyer2008-anm, peters14a-resit-paper}, we investigate the extent to which these guarantees translate to the latent setting.

We tackle the challenge of unsupervised identifiability by operationalizing the ANM constraint on the latent factors. Theoretically, we demonstrate that this constraint resolves component-wise indeterminacy by reducing the admissible diffeomorphism to the affine class (Theorem~\ref{thm:linearization}), providing a tractable optimization target even though adversarial mixing remains a theoretical possibility (Proposition~\ref{prop:mixing_impossibility}). Methodologically, we propose LANCA, which utilizes a deterministic Wasserstein Auto-Encoder (WAE) \citep{Tolstikhin2017WassersteinA} and a differentiable ANM layer. By replacing stochastic sampling with deterministic encoding, LANCA explicitly optimizes residual independence, allowing the model to recover latent causal factors directly from observational data. Empirically, we show that LANCA outperforms existing baselines across synthetic physics simulations and photorealistic benchmarks.

\section{Related Work}
\label{sec:related_work}

\subsection{Disentangled Representation Learning and the Limits of Independence}
The dominant paradigm in DRL equates disentanglement with statistical independence, exemplified by VAE variants like $\beta$-VAE \citep{higgins2017betavae} and FactorVAE \citep{kim18b-factorvae}. However, statistical independence is insufficient to uniquely identify factors in the general nonlinear setting. \citet{locatello19a-assumptions} proved that without inductive biases, unsupervised disentanglement is theoretically impossible, as the true joint distribution is indistinguishable from entangled mixtures. Consequently, an unsupervised model optimizing solely for independence cannot distinguish between the true factors $\mathbf{s}$ and an entangled representation $\mathbf{z} = \boldsymbol{\Psi}(\mathbf{s})$, for some general nonlinear mixing function $\boldsymbol{\Psi}$.

\subsection{Causal Representation Learning}

To address the limitations of DRL, CRL assumes that latent factors are causally related rather than being mutually independent \citep{Scholkopf2021TowardsCR}. CRL argues that the independence assumption can be invalid since real-world latent factors are often causally related, and enforcing independence actively suppresses the causal structure. However, CRL also inherits the fundamental challenge of identifiability. Without additional constraints, the observational distribution $\mathrm{P}(\mathbf{x})$ is insufficient to distinguish the true causal model from entangled alternatives. Existing solutions typically rely on auxiliary variables derived from non-i.i.d. or multi-environment signals. For example, methods like iVAE \citep{Khemakhem2019iVAE} and TCL \citep{Hyvrinen2016TCL} exploit temporal correlations or environmental indices as proxies for interventions that render the latent factors identifiable.

In the static i.i.d. (i.e., observational) setting, where such auxiliary signals are unavailable, recent methods have attempted to integrate SCMs directly into the generative process. CausalVAE \citep{Yang2020CausalVAESC} and DEAR \citep{Shen2020WeaklySD} successfully model causal dependencies but bypass the unsupervised identifiability problem by relying on \textit{supervision}. CausalVAE requires ground-truth concept labels to condition the prior, while DEAR relies on a weakly supervised setting involving labels or graph skeletons. Thus, learning identifiable causal representations from i.i.d. data in a fully unsupervised manner remains an open challenge. LANCA addresses this gap by replacing supervision with an ANM constraint, exploring the extent to which structural assumptions can substitute for labels in recovering latent causal factors.

\subsection{Identifiability via Additive Noise Models}
LANCA draws inspiration from causal discovery, where \citet{hoyer2008-anm} and \citet{peters14a-resit-paper} demonstrated that structural equations of the form $s_i = h_i(\mathbf{pa}_i) + n_i$ (with nonlinear $h_i$ and independent $n_i$) render the causal graph identifiable from observational data. While full identifiability holds for observed data, learning latent factors introduces complexities. \citet{Welch2024IdentifiabilityGF} established that for latent ANMs under \textit{linear generative mechanisms} $\mathbf{x} = \mathbf{H}\mathbf{z}$, identifiability is limited to ancestral layers. We extend this analysis to the more challenging domain of \textit{nonlinear generative mechanisms}, $\mathbf{x} = \mathbf{g}(\mathbf{z})$, which corresponds to LANCA's decoder. This setting remains largely unexplored, as standard ANM guarantees do not automatically account for interactions between structural equations and complex generative mappings.

\section{Identifiability Analysis}
\label{sec:theory}

In this section, we analyze the theoretical limits of identifying latent causal factors from observational data. We specifically investigate the restrictive capacities of the ANM constraint under two distinct forms of entanglement: \textit{component-wise nonlinear distortion} and \textit{general nonlinear mixing}.

Our analysis suggests that while the ANM constraint is sufficient to recover endogenous variables, i.e., variables with parents, up to linear transformations against component-wise distortions, it does not theoretically guarantee uniqueness against general nonlinear mixing.

\subsection{Identifiability under Component-Wise Distortion}
Consider the case where the learned latent variables $z_i$ are component-wise nonlinear transformations of the true factors $s_i$, such that $z_i = \psi_i(s_i)$. We assume the model learns a representation $\mathbf{z}$ that satisfies the ANM structural equation:
\begin{equation}
    \label{eq:learned_anm}
    z_i = f_i(\widehat{\mathbf{pa}}_i) + \epsilon_i, \quad \text{where } \epsilon_i \indep \widehat{\mathbf{pa}}_i, \quad \forall i.
\end{equation}
Under the standard assumptions of differentiability (Assumption~\ref{ass:differentiability}) and structural consistency (Assumption~\ref{ass:structural_consistency}), we show that strictly enforcing this structural requirement effectively restricts the admissible transformations $\psi_i$ to the affine class.

\begin{theorem}[Linearization of Endogenous Variables]
\label{thm:linearization}
Let $s_i$ be an endogenous ground-truth variable (i.e., $\mathbf{pa}_i \neq \emptyset$) generated by a valid ANM $s_i = h_i(\mathbf{pa}_i) + n_i$, where $h_i$ is a non-constant, differentiable function of the parents $\mathbf{pa}_i$. Let $z_i = \psi_i(s_i)$ be a smooth, invertible transformation of $s_i$. If the transformed variable $z_i$ constitutes a valid ANM satisfying Eq.~\ref{eq:learned_anm}, then $\psi_i$ must be an affine transformation.
\end{theorem}

This theorem (proven in Appendix~\ref{app:proof_theorem1}) resolves the component-wise indeterminacy inherent in nonlinear ICA \citep{hyvarinen1999-nonlinearICA-analysis}. While statistical independence is typically invariant under arbitrary component-wise nonlinear transformations, the ANM constraint effectively ``linearizes'' the representation for endogenous variables, reducing the admissible space of diffeomorphisms to simple affine transformations.

\subsection{Impossibility Result for General Nonlinear Mixing}
A stronger form of entanglement involves \textit{mixing} multiple true factors, i.e., $\mathbf{z} = \boldsymbol{\Psi}(\mathbf{s})$. Ideally, we seek a guarantee that the ANM constraint rejects such mixtures. However, we demonstrate that ANMs are susceptible to a specific form of mixing where the encoder negates the structural equation.
\setcounter{theorem}{0}
\begin{proposition}[Existence of Spurious Independent Solutions]
\label{prop:mixing_impossibility}
Let the true data generating process be $s_1 = n_1$ and $s_2 = h_2(s_1) + n_2$, where $n_1 \indep n_2$. There exists a smooth, invertible mixing function $\boldsymbol{\Psi}: \mathcal{S} \to \mathcal{Z}$ such that the resulting variables $z_1, z_2$ appear to be mutually independent exogenous variables, thereby erasing the true causal structure while satisfying the ANM independence objective.
\end{proposition}

We show by the counter-example in Appendix~\ref{app:proof_prop1} that if the mixing function $\boldsymbol{\Psi}$ is sufficiently expressive, it can ``absorb'' the structural mechanism. Consequently, \textit{observational} identifiability under general nonlinear mixing is formally impossible without restricting the functional class of the encoder. This result underscores the severity of the challenge: the mathematical solution space includes entangled representations that statistically mimic the true model. LANCA accepts this theoretical boundary but demonstrates that it is not an empirical dead end. In our experiments, we observe that LANCA tends to recover the underlying causal structure, suggesting that these complex adversarial inversions, while theoretically valid, are not the solutions found by the optimization process in practice.

\section{Methodology}
\label{sec:method}

\begin{figure}[!ht]
    \centering
    \includegraphics[width=\linewidth]{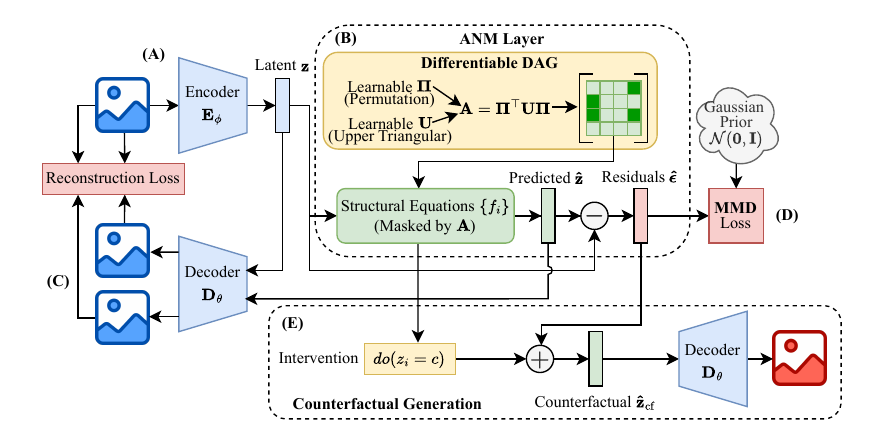}
    \caption{\textbf{The Latent ANM Causal Autoencoder (LANCA).} (A) The encoder $\mathbf{E}_\phi$ maps $\mathbf{x}$ to a deterministic latent code $\mathbf{z}$ via the WAE framework. (B) The \textbf{ANM Layer} learns the adjacency matrix $\mathbf{A} = \mathbf{\Pi}^\top \mathbf{U} \mathbf{\Pi}$, the set of structural equations $\{f_i\}_{i=1}^n$, and performs structural abduction to recover the residual vector $\boldsymbol{\hat{\epsilon}}$. (C) The objective enforces reconstruction accuracy on both the direct encoding path and the ANM-regenerated path. (D) The model matches the joint distribution of residuals $\mathrm{Q}(\boldsymbol{\hat{\epsilon}})$ to a factorized prior $\mathrm{P}(\boldsymbol{\epsilon})$ via Maximum Mean Discrepancy (MMD). (E) Counterfactual generation is achieved via the three-step procedure: abduction of $\boldsymbol{\hat{\epsilon}}$, intervention on specific factors $do(z_i=c)$ for some constant $c$, and prediction of the new observation $\mathbf{x}_\text{cf}=\mathbf{D}_\theta(\mathbf{z}_\text{cf})$.}
    \label{fig:lanca_overview}
\end{figure}

Guided by the theoretical properties discussed in \cref{sec:theory}, we introduce a practical implementation of LANCA. While we acknowledge the theoretical boundaries of observational identifiability, we posit that explicit architectural constraints can serve as a strong inductive bias, steering the optimization away from adversarial entanglements and towards the true causal structure. Our framework, illustrated in \cref{fig:lanca_overview}, implements two fundamental design choices to operationalize this bias: a \textbf{deterministic encoding} scheme to precisely isolate the ANM's residuals, and a \textbf{differentiable ANM layer} capable of learning both the structural equations and the underlying causal graph.

\subsection{Deterministic Encoding via Wasserstein Auto-Encoders}
We assume the observed data $\mathbf{x} \in \mathbb{R}^D$ is generated by latent factors $\mathbf{z} \in \mathbb{R}^n$ via an injective projection. While our theoretical analysis adopts the standard manifold assumption (implying a diffeomorphism), in practice, the ambient dimension of the observation space is significantly larger than the latent dimension ($D \gg n$). Consequently, the encoder $\mathbf{E}_\phi: \mathbb{R}^D \to \mathbb{R}^n$ acts as an approximate injective mapping designed to invert this mechanism specifically over the support of the data distribution.

A critical challenge in learning latent causal factors is the precise recovery of the ANM residuals. Standard VAEs model the encoder $q(\mathbf{z}|\mathbf{x})$ stochastically, typically as a factorized Gaussian. This introduces a fundamental ambiguity: if $\mathbf{z}$ is sampled stochastically, the structural residual $\hat{\epsilon}_i = z_i - f_i(\widehat{\mathbf{pa}}_i)$ becomes a mixture of the true exogenous noise and the variance injected by the encoder. This stochasticity prevents the precise independence optimization required for identifiability. 

To resolve this, we adopt the WAE framework \citep{Tolstikhin2017WassersteinA}. Unlike VAEs, which enforce the evidence lower bound (ELBO) constraint on every individual sample, WAEs minimize the divergence between the \textit{aggregated} posterior $\mathrm{Q}(\mathbf{z}) = \mathbb{E}_{\mathrm{P}(\mathbf{x})}[q(\mathbf{z}|\mathbf{x})]$ and the prior. As shown by \citet{Tolstikhin2017WassersteinA}, this relaxation removes the requirement for non-degenerate Gaussian encoders, allowing the conditional distribution $q(\mathbf{z}|\mathbf{x})$ to be modeled as a deterministic Dirac delta function $\mathrm{Q}(\mathbf{z}|\mathbf{x})=\delta_{\boldsymbol{\mu}(\mathbf{x})}$. We therefore define the encoder as a deterministic map $\mathbf{z} = \mathbf{E}_\phi(\mathbf{x})$.

\subsection{The Differentiable ANM Layer}
At the core of LANCA is the \textbf{ANM Layer}, which imposes an ANM structure on the latent space. We assume that $\mathbf{z}$ follows the structural equation defined in \cref{eq:learned_anm}. We implement the structural functions $f_i$ as nonlinear mechanisms parameterized by neural networks, and the residuals $\epsilon_i$ are trained to be mutually independent exogenous noise terms. We aim to learn the adjacency matrix $\mathbf{A} \in \{0,1\}^{n \times n}$ and the functions $\{f_i\}_{i=1}^n$ simultaneously.

\paragraph{Differentiable DAG via Matrix Factorization.}
To enforce the acyclicity constraint efficiently, we adopt the matrix factorization approach proposed by \citet{Charpentier2022DifferentiableDS}, parameterizing the adjacency matrix as $\mathbf{A} = \mathbf{\Pi}^\top \mathbf{U} \mathbf{\Pi}$. In this formulation, $\mathbf{\Pi}$ is a permutation matrix encoding the topological order, and $\mathbf{U}$ is a strictly upper-triangular matrix of edge weights, a factorization that guarantees any derived $\mathbf{A}$ is a DAG. 
We adapt the probabilistic formulation of \citet{Charpentier2022DifferentiableDS} to our deterministic setting. We use the \textsc{SoftSort} operator \citep{Prillo2020SoftSortAC} for permutation learning, the sigmoid function for the edge weights, and the straight-through estimator (STE) \citep{Bengio2013StraightThrough} to propagate gradients through the discrete graph structure.


\paragraph{Mechanism Learning and Structural Abduction.}

The structural equations are modeled using a set of multi-layer perceptrons (MLPs), with a separate network parameterizing each equation. Consistent with the standard definition of ANMs \citep{hoyer2008-anm} and our theoretical analysis—which relies on differentiability to establish the linearization of endogenous variables (Theorem~\ref{thm:linearization})—we employ smooth activation functions (e.g., SiLU, Tanh). During the forward pass, we perform structural abduction to estimate the exogenous noise. Given the latent factors $\mathbf{z}$, we compute the predicted value $\hat{z}_i = f_i(\mathbf{z} \circ \mathbf{A}_{:,i})$ and recover the residual $\hat{\epsilon}_i = z_i - \hat{z}_i$. This contrasts with prior approaches like CausalVAE \citep{Yang2020CausalVAESC}, which first learns independent residuals via a VAE before generating $\mathbf{z}$. By calculating residuals from deterministically encoded latents, we transform the residual independence from an \textit{assumption} to an \textit{optimizable objective}, thereby enforcing the ANM constraints to encourage structural disentanglement.

\subsection{Optimization Objective}
The training objective is a weighted sum of three components: reconstruction fidelity, causal independence, and structural regularization.

\paragraph{Structural Consistency Reconstruction ($\mathcal{L}_{\text{recon}}$).} 
We minimize the reconstruction error of the observed data $\mathbf{x}$. Crucially, to enforce that the ANM accurately captures the generative process, we require the decoder $\mathbf{D}_\theta$ to reconstruct the input $\mathbf{x}$ from \textit{both} the deterministically encoded latent $\mathbf{z}$ and the structurally re-generated latent $\mathbf{z}_{\text{scm}}$. The loss is given by:
\begin{equation}
    \mathcal{L}_{\text{recon}} = \|\mathbf{x} - \mathbf{D}_\theta(\mathbf{z})\|^2_2 + \lambda_{\text{scm}} \|\mathbf{x} - \mathbf{D}_\theta(\mathbf{z}_{\text{scm}})\|^2_2.
\end{equation}

\paragraph{Independence via MMD ($\mathcal{L}_{\text{indep}}$).}
Our identifiability framework relies on the residuals $\boldsymbol{\epsilon} = [\epsilon_1, \dots, \epsilon_n]^\top$ being mutually independent. We enforce this by minimizing the Maximum Mean Discrepancy (MMD) \citep{gretton2012MMD} between the joint distribution of the abducted residuals $\mathrm{Q}(\boldsymbol{\hat{\epsilon}})$ and a factorized prior $\mathrm{P}(\boldsymbol{\epsilon}) = \mathcal{N}(\mathbf{0}, \mathbf{I})$. To robustly capture distributional discrepancies across different modes, we employ a multi-scale kernel strategy, denoted by $k(\cdot,\cdot)$, defined as a sum of base kernels with varying bandwidths. While the Radial Basis Function (RBF) is standard, \citet{Tolstikhin2017WassersteinA} argue that the Inverse Multiquadratic (IMQ) kernel often yields superior performance in high-dimensional matching due to its heavier tails, which prevent gradient vanishing on outliers. Consequently, to ensure robustness, we treat the kernel type (RBF vs. IMQ) as a tunable hyperparameter in our optimization process. The independence loss is computed as $\mathcal{L}_{\text{indep}} = \text{MMD}_k(\mathrm{Q}(\boldsymbol{\hat{\epsilon}}), \mathrm{P}(\boldsymbol{\epsilon}))$.

\paragraph{Structural Regularization.}
While our matrix factorization ensures acyclicity by design, we impose additional regularization. First, to encourage parsimony, we apply a sparsity constraint $\mathcal{L}_{\text{sparse}}$ formulated as the binary cross entropy (BCE) between the learned edge probabilities and a fixed low-sparsity prior (e.g., $p=0.01$), which we tune. This suppresses dense connections in $\mathbf{U}$.
Second, to enforce discreteness, we minimize the entropy $\mathcal{L}_{\text{ent}}$ of the soft permutation matrix $\mathbf{\Pi}$. This penalizes uncertainty in the ordering, forcing $\mathbf{\Pi}$ to converge towards a hard permutation matrix. 

\paragraph{Total Objective.}
The final training objective aggregates the reconstruction fidelity, independence, and structural regularization terms:
\begin{equation}
    \mathcal{L} = \mathcal{L}_{\text{recon}} + \beta \mathcal{L}_{\text{indep}} + \gamma_1 \mathcal{L}_{\text{sparse}} + \gamma_2 \mathcal{L}_{\text{ent}}.
    \label{eq:total_loss}
\end{equation}

\subsection{Optimization Strategy}
Simultaneously optimizing the adjacency matrix $\mathbf{A}$ and the functional parameters, i.e., encoder $\phi$, decoder $\theta$, and our nonlinear structural equations $\{f_i\}$, is notoriously unstable. As observed by \citet{Yang2020CausalVAESC} even in linear settings, this joint optimization often results in premature convergence to empty graphs. To mitigate this, we employ a three-pronged schedule (more details in Appendix~\ref{app:optimization_strategy}). First, via \textbf{differential learning rates}, we disentangle the learning dynamics by assigning distinct rates to the functional parameters, edge weights $\mathbf{U}$, and permutation scores $\mathbf{\Pi}$. Empirically, prioritizing the permutation scores ($\eta_{\text{perm}} > \eta_{\text{edge}}$) encourages the model to resolve the global topological ordering before fine-tuning specific edge dependencies. Second, we apply \textbf{temperature annealing} to the edge temperature $\tau_{\text{edges}}$, transitioning from soft, high-variance gradients to hard, discrete edges to enable exploration early in training. Finally, we implement a \textbf{sparsity warmup}, holding the regularization weight $\gamma_1$ at zero during the initial phase to prevent the strong sparsity prior from suppressing edges before meaningful causal mechanisms are learned.


\subsection{Counterfactual Generation}
Once trained, LANCA enables counterfactual generation via the standard three-step counterfactual estimation procedure: \textit{abduction, action, and prediction} \citep{pearl2009causality}. Given an observation $\mathbf{x}$, we first \textbf{abduct} the exogenous noise by encoding the latent representation $\mathbf{z} = \mathbf{E}_\phi(\mathbf{x})$ and recovering the residuals $\hat{\epsilon}_i = z_i - f_i(\widehat{\mathbf{pa}}_i)$ for each causal variable $i$. In the \textbf{action} step, we specify an intervention $do(z_i = c)$ for some constant $c$. Finally, in the \textbf{prediction} step, we construct the counterfactual latent $\mathbf{z}_{\text{cf}}$ by traversing the causal graph in topological order: for the intervened node, we set $z_{\text{cf}, i} = c$; for all other nodes, we recompute $z_{\text{cf}, i} = f_j(\widehat{\mathbf{pa}}_{\text{cf}, i}) + \hat{\epsilon}_i$ using the learned mechanisms and the abducted noise. The counterfactual image is then generated via the decoder $\mathbf{x}_{\text{cf}} = \mathbf{D}_\theta(\mathbf{z}_{\text{cf}})$.

\section{Experiments}

\subsection{Datasets}

\paragraph{Synthetic Physics Benchmarks: Pendulum and Flow.} We adopt the \textit{Pendulum} and \textit{Flow} environments introduced by \citet{Yang2020CausalVAESC}. In \textbf{Pendulum}, the \textit{Pendulum Angle} and \textit{Light Position} serve as exogenous variables that causally determine the endogenous \textit{Shadow Length} and \textit{Shadow Position}. The \textbf{Flow} dataset simulates fluid dynamics where \textit{Ball Size} and \textit{Hole Position} influence \textit{Water Height} and the resulting \textit{Water Flow} via Archimedes' principle and Torricelli's law. To construct a testbed consistent with the ANM assumption, we inject Gaussian noise, $\epsilon \sim \mathcal{N}(0, (\sigma_{v} \cdot \eta)^2)$, into the endogenous variables. We scale the noise variance by the variable's standard deviation $\sigma_{v}$ (with $\eta = 0.1$) to ensure the noise is proportional to the signal's dynamic range. The final datasets consist of 5,899 training and 1,409 test samples for \textit{Pendulum}, and 6,533 training and 1,567 test samples for \textit{Flow}.

\paragraph{CANDLE.} We also use the photorealistic CANDLE dataset \citep{Reddy2021CANDLE}, which renders 3D objects within complex high-dynamic range image (HDRI) backgrounds. The data is governed by a two-level causal graph containing six interpretable factors (\textit{Light, Scene, Object, Size, Color, Rotation}). Crucially, while these factors do not causally influence each other, they are mutually confounded by background scene constraints (e.g., large objects cannot be rendered in some indoor scenes because they cannot fit), creating strong spurious correlations. The dataset consists of 10,664 training and 1,882 test samples. 
We resize images to $96 \times 96$ and use the provided bounding boxes to apply a localized reconstruction loss $\mathcal{L}_{bbox}$ for all models, ensuring small foreground objects are encoded.

\subsection{Evaluation Metrics}

\paragraph{Standard Disentanglement Metrics.}
For completeness, we report the Mutual Information Gap (MIG) \citep{Chen2018MIG}, a standard metric in DRL literature that quantifies the compactness of latent representations.

\paragraph{Graph Structure and Alignment (Pendulum \& Flow).}
For the synthetic physics benchmarks, we evaluate structural recovery using Structural Hamming Distance (SHD) and Structural Intervention Distance (SID) \citep{Peters2015SID}. To handle the permutation ambiguity of unsupervised methods, we align learned latents to ground-truth factors by maximizing mutual information via the Hungarian algorithm \citep{Kuhn1955Hungarian}. We further report the Mean Mutual Information (MMI) of the matched pairs to quantify how well the latent factors capture the true generative factors. We explicitly exclude robustness metrics like the Interventional Robustness Score (IRS) \citep{Sutter2019IRS} from these environments; since physical systems inherently involve direct causal dependencies, intervening on a cause \textit{must} change the representation of its effect. Applying robustness metrics here would incorrectly penalize the model for learning the true physical mechanisms.

\paragraph{Interventional Robustness and Confounding (CANDLE).}
In contrast to the physics benchmarks, the \textit{CANDLE} dataset's ground truth factors do not possess direct causal links but are mutually confounded by the scene context. Consequently, standard graph recovery metrics like SHD and SID are not suitable, as there are no direct edges between endogenous factors to recover. Instead, we evaluate the disentanglement of these confounded factors using the IRS, which measures the stability of a feature representation when other factors are intervened upon while the target factor is held fixed. Complementary to this, we report \textbf{Unconfoundedness (UC)} and \textbf{Counterfactual Generativeness (CG)} \citep{Reddy2021CANDLE}. UC quantifies the structural separation of mechanisms by measuring the extent to which distinct generative factors are mapped to unique, non-overlapping latent subspaces; high overlap implies the model has absorbed the confounding correlations rather than isolating the factors. CG evaluates the causal validity of the generation: it measures whether intervening on a specific latent factor produces the expected semantic change in the image (e.g., changing object size) without inadvertently altering other attributes (e.g., the background). These metrics are specifically tailored to quantify robustness against spurious correlations, making them well-suited for the confounding structure of CANDLE.

\subsection{Experimental Setup}
\paragraph{Baselines.} We compare LANCA against: (1) \textbf{$\beta$-VAE} and \textbf{FactorVAE} (DRL baselines); and (2) \textbf{CausalVAE} \citep{Yang2020CausalVAESC} in both supervised and unsupervised (label-ablated) settings. All models utilize identical encoder/decoder backbones (details in Appendix~\ref{app:architecture}) to isolate the contribution of the structural learning layers.

\paragraph{Implementation and Model Selection.}
For physics environments, we set $z_\text{dim}=4$. For CANDLE, we set $z_\text{dim}=12$ to allow sufficient capacity to absorb confounders. Acknowledging the difficulty of unsupervised model selection \citep{locatello19a-assumptions}, we employ an ``oracle validation'' strategy where hyperparameters are selected based on ground-truth metrics: MIG (independence models) and MMI (causal models) for physics datasets, and UC ($\rho=5$) for all models on CANDLE. Final results are averaged over 10 seeds; the full model selection process is detailed in Appendix~\ref{app:tuning}.

\section{Results and Discussion}

\subsection{Unsupervised Causal Structure Learning}

A primary contribution of LANCA is the empirical ability to discover underlying causal graphs without supervision. As reported in Table~\ref{tab:results_pendulum_flow}, LANCA achieves the best graph recovery error rates among the causal methods across both physics benchmarks.

On the \textit{Pendulum} dataset, LANCA attains an SHD of $\mathbf{4.0}$ and SID of $\mathbf{5.5}$, outperforming the unsupervised CausalVAE (SHD $5.3$, SID $7.6$). This performance is competitive even with the \textit{supervised} CausalVAE (SHD $7.4$, SID $8.0$), suggesting that our structural constraints are effective substitutes for label supervision in simple environments. 
We hypothesize that the supervised CausalVAE may have struggled with the stochasticity of our modified data generating process (Gaussian noise injection), whereas LANCA's deterministic encoding effectively separates this noise from the structural mechanism.

The strength of LANCA is most evident in the complex fluid dynamics task (\textit{Flow}). Here, LANCA achieves a Mean Mutual Information (MMI) of $\mathbf{1.120}$, outperforming the unsupervised CausalVAE ($0.381$). This massive gap highlights a critical limitation in prior work: while CausalVAE assumes a Linear SCM in the latent space, physical systems like fluid dynamics are inherently nonlinear. LANCA's use of nonlinear mechanisms allows it to capture these dynamics accurately, a capability further reflected in it achieving the lowest SHD ($\mathbf{4.3}$) among all models, including the supervised baseline. 
Finally, LANCA yields lower MIG scores which is expected as MIG rewards statistical independence, whereas LANCA explicitly models causal dependencies.

\begin{table*}[!ht]
\centering
\caption{\textbf{Quantitative results on synthetic physics environments (Pendulum \& Flow).} Mean $\pm$ std over 10 seeds. LANCA consistently achieves the lowest structural errors (SHD, SID) across both environments and good latent factor alignment (MMI) on the complex Flow dataset.}
\label{tab:results_pendulum_flow}
\resizebox{\textwidth}{!}{
\begingroup
\setlength{\aboverulesep}{0pt}
\setlength{\belowrulesep}{0pt}
\renewcommand{\arraystretch}{1.2}
\begin{tabular}{l|cccc|cccc}
\toprule
 \multirow{2}{*}{\textbf{Model}} & & \multicolumn{2}{c}{\textbf{Pendulum}} & & & \multicolumn{2}{c}{\textbf{Flow}} & \\
 & \textbf{MIG} $\uparrow$ & \textbf{MMI} $\uparrow$ & \textbf{SHD} $\downarrow$ & \textbf{SID} $\downarrow$ & \textbf{MIG} $\uparrow$ & \textbf{MMI} $\uparrow$ & \textbf{SHD} $\downarrow$ & \textbf{SID} $\downarrow$ \\
\midrule
$\beta$-VAE & 0.202 $\pm$ 0.08 & 0.915 $\pm$ 0.17 & - & - & 0.006 $\pm$ 0.00 & 0.018 $\pm$ 0.01 & - & - \\
Factor-VAE & 0.311 $\pm$ 0.17 & 1.210 $\pm$ 0.31 & - & - & 0.421 $\pm$ 0.16 & 0.432 $\pm$ 0.16 & - & - \\
\midrule
CausalVAE (Sup) & \textit{\textbf{0.585 $\pm$ 0.08}} & \textit{\textbf{2.006 $\pm$ 0.17}} & \textit{7.400 $\pm$ 0.52} & \textit{8.000 $\pm$ 0.00} & \textit{\textbf{0.770 $\pm$ 0.03}} & \textit{\textbf{2.221 $\pm$ 0.08}} & \textit{5.800 $\pm$ 1.14} & \textit{8.000 $\pm$ 0.00} \\
CausalVAE (Unsup) & 0.201 $\pm$ 0.04 & 1.006 $\pm$ 0.15 & 5.300 $\pm$ 1.49 & 7.600 $\pm$ 1.71 & 0.224 $\pm$ 0.11 & 0.381 $\pm$ 0.17 & 4.900 $\pm$ 1.20 & 6.900 $\pm$ 2.28 \\
LANCA (Ours) & 0.172 $\pm$ 0.07 & 0.997 $\pm$ 0.21 & \textbf{4.000 $\pm$ 0.82} & \textbf{5.500 $\pm$ 1.51} & 0.015 $\pm$ 0.02 & 1.120 $\pm$ 0.03 & \textbf{4.300 $\pm$ 1.06} & \textbf{5.300 $\pm$ 2.54} \\
\bottomrule
\end{tabular}

\endgroup
}
\end{table*}

\subsection{Interventional Robustness Under Confounding}

\paragraph{Robustness to Spurious Correlations (IRS).}
As shown in Table~\ref{tab:results_candle}, LANCA achieves a dominant IRS of $\mathbf{0.800}$, outperforming all baselines by a large margin. This result indicates that standard baselines essentially act as ``sponges'' for spurious correlations, absorbing complex HDRI background features (e.g., lighting changes) into the object representations. In contrast, LANCA explicitly models the structural equations, allowing it to isolate the target object properties from the confounding scene context. This high IRS confirms that LANCA has learned representations that remain stable against changes in other generative factors, satisfying the core desiderata of causal disentanglement.

\paragraph{Unconfoundedness and Generative Capabilities.}
LANCA achieves the highest UC score ($\mathbf{0.518}$), significantly outperforming Factor-VAE ($0.270$) and CausalVAE ($0.429$). This result validates our over-parameterization strategy ($z_{dim}=12$ for 6 factors). By learning extra latents, LANCA can allocate separate latent factors to model the confounding mechanisms themselves, leaving the subspaces corresponding to the true target factors unconfounded. 

Regarding generative quality, while the CG scores are generally low across all methods due to the complexity of the dataset, LANCA remains competitive with the best baselines (CG $\rho=1$: $0.061$ vs $\beta$-VAE: $0.062$). This suggests that LANCA does not compromise generative fidelity to achieve its superior structural disentanglement. It successfully balances the trade-off, offering robust, unconfounded features (high UC/IRS) while maintaining a generative capacity on par with the baselines.

\begin{table*}[!ht]
\centering
\caption{\textbf{Results on the photorealistic CANDLE dataset.} Mean $\pm$ std over 10 seeds. LANCA achieves the best IRS and UC, verifying its ability to separate causal factors from the background scene confounders.}
\label{tab:results_candle}
\resizebox{\textwidth}{!}{
\begin{tabular}{lcccccc}
\toprule
\textbf{Model} & \textbf{MIG} $\uparrow$ &  \textbf{IRS} $\uparrow$ & \textbf{UC ($\rho=1$)} $\uparrow$ & \textbf{CG ($\rho=1$)} $\uparrow$ & \textbf{UC ($\rho=5$)} $\uparrow$ & \textbf{CG ($\rho=5$)} $\uparrow$ \\
\midrule
$\beta$-VAE & 0.019 $\pm$ 0.01 &  0.440 $\pm$ 0.01 & 0.387 $\pm$ 0.17 & \textbf{0.062 $\pm$ 0.01} & 0.401 $\pm$ 0.05 & \textbf{0.088 $\pm$ 0.02} \\
Factor-VAE & 0.019 $\pm$ 0.01 &  0.401 $\pm$ 0.02 & 0.400 $\pm$ 0.19 & 0.048 $\pm$ 0.01 & 0.270 $\pm$ 0.12 & 0.062 $\pm$ 0.02 \\
\midrule
CausalVAE (Unsup) & \textbf{0.031 $\pm$ 0.01} & 0.452 $\pm$ 0.01 & 0.520 $\pm$ 0.18 & 0.054 $\pm$ 0.02 & 0.429 $\pm$ 0.08 & 0.027 $\pm$ 0.01 \\
LANCA (Ours) & 0.017 $\pm$ 0.01 & \textbf{0.800 $\pm$ 0.03} & \textbf{0.560 $\pm$ 0.13} & 0.061 $\pm$ 0.02 & \textbf{0.518 $\pm$ 0.09} & 0.037 $\pm$ 0.02 \\
\bottomrule
\end{tabular}
}
\end{table*}

\subsection{Qualitative Analysis}

\begin{figure*}[t]
    \centering
    \includegraphics[width=\textwidth]{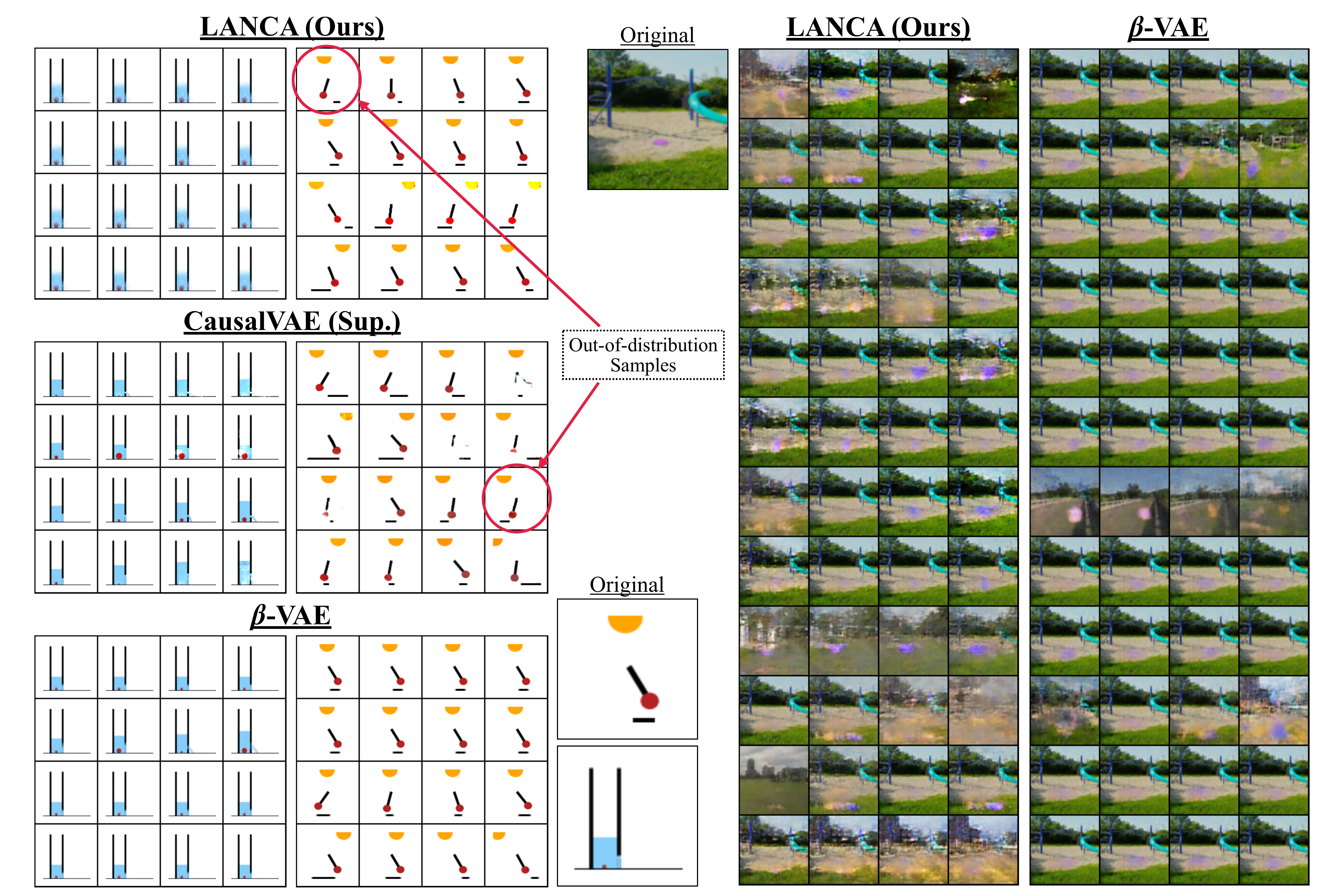}
    \caption{\textbf{Qualitative Counterfactual Generation.} Comparison of counterfactual images generated by LANCA (Ours), CausalVAE (Supervised), and $\beta$-VAE.}
    \label{fig:qualitative}
\end{figure*}

We qualitatively assess the learned causal disentanglement by inspecting counterfactual images generated via the do-operator (Figure~\ref{fig:qualitative}). To generate these samples, we intervene on a single latent factor $z_i$ by setting it to a fixed value $c$ (action). For causal models (LANCA, CausalVAE), this intervention propagates to downstream effects via the learned SCM (prediction) before decoding; for the non-causal baseline ($\beta$-VAE), we simply traverse the target latent dimension. In Figure~\ref{fig:qualitative}, each row depicts the decoding results as we sweep a single factor across four distinct values.

\paragraph{Causal Consistency.}
An indication of a correct causal model is the asymmetry of intervention: intervening on a cause should influence its effects, while intervening on an effect should leave the cause invariant \citep{Yang2020CausalVAESC}. On the \textit{Pendulum} dataset, LANCA demonstrates this asymmetry. When the \textit{Light Source} (cause) is varied, the shadow changes appropriately, but the \textit{Pendulum Angle} remains fixed (row 4 of LANCA). Crucially, as highlighted by the red circles, intervening on the effect (e.g., fixing the angle but moving the shadow) results in ``impossible'' out-of-distribution images. The ability to generate these physically invalid but causally consistent states confirms that LANCA and CausalVAE have learned the underlying mechanism rather than merely memorizing the joint distribution—a capability absent in the $\beta$-VAE baseline, which generates generic in-distribution samples regardless of the intervention.

\paragraph{Disentangling Objects from Backgrounds.}
On \textit{CANDLE}, qualitative results visually confirm the high UC scores reported in Table~\ref{tab:results_candle}. Despite the complexity of the HDRI backgrounds, LANCA successfully disentangles the foreground object from the scene context. As shown in the 9th row (LANCA), the model is able to intervene on the background scene while preserving the foreground object's properties. Conversely, LANCA was able to vary the object size while keeping the complex background fixed (e.g., rows 3, 5, 6, and 7). These mirror our quantitative findings: the model has successfully broken the spurious dependence between the object and its background scene.

\section{Conclusion}
\label{sec:conclusion}

In this work, we introduced LANCA, a framework that effectively uses the ANM constraint to recover latent causal structure from observational data. While our theoretical analysis confirms that latent ANMs do not guarantee uniqueness against general nonlinear mixing, we demonstrate that the ANM assumption serves as a powerful inductive bias, enabling robust causal structure recovery in the \textit{Pendulum} and \textit{Flow} datasets and superior IRS and UC on the challenging \textit{CANDLE} dataset.

Crucially, LANCA resolves key design limitations in prior state-of-the-art methods like CausalVAE \citep{Yang2020CausalVAESC}. By modeling nonlinear causal mechanisms, we capture complex physical dynamics (e.g., fluid flow) that linear models miss. By replacing the stochastic encoder with a deterministic WAE framework, we transform residual independence from a passive assumption into an explicit MMD optimization objective. Finally, our adaptation of differentiable matrix factorization \citep{Charpentier2022DifferentiableDS} to the discrete domain guarantees acyclicity by construction. This avoids the instability inherent in the NOTEARS-based approach used in CausalVAE, which relies on an expensive matrix exponential trace constraint optimized via Augmented Lagrangian that penalizes cycles rather than structurally precluding them.

\section{Limitations and Future Work}
\label{sec:limitations}

Despite these advances, our approach operates within strict boundaries. Theoretically, the ANM constraint alone cannot guarantee uniqueness against general nonlinear mixing in the observational setting; full provable identifiability likely requires non-i.i.d. data or weak supervision. Methodologically, our framework assumes additive noise and acyclicity, limiting its immediate applicability to systems with multiplicative noise or feedback loops. Nevertheless, this work represents a first step towards bridging classical causal discovery and deep representation learning by operationalizing structural constraints as inductive biases. Future work can extend this foundation by exploring broader identifiable classes, such as Post-Nonlinear Models, or by integrating LANCA with auxiliary signals—such as non-i.i.d. data or weak supervision—leveraging our structural constraints to enhance identifiability where data-driven signals alone may be insufficient.

\bibliography{refs.bib}
\appendix

\section{Detailed Proofs}
\label{app:proofs}

\paragraph{Notation and Assumptions.}
We rely on the following regularity conditions:

\begin{assumption} [Differentiability] 
\label{ass:differentiability}
The true latent generative mechanism $s_i = h_i(\mathbf{pa}_i) + n_i$ is differentiable and non-trivial, i.e., $\nabla_{\mathbf{pa}_i} h_i(\mathbf{pa}_i) \neq \mathbf{0}$ almost everywhere. The transformation $z_i = \psi_i(s_i)$ is a smooth diffeomorphism.
\end{assumption} 

\begin{assumption} [Structural Consistency] 
\label{ass:structural_consistency}
We assume the learner has identified the correct topological causal structure. Consequently, the learned parents $\widehat{\mathbf{pa}}_i$ correspond to the component-wise transformations of the true parents, i.e., $\widehat{\mathbf{pa}}_i = \boldsymbol{\psi}_{\mathbf{pa}_i}(\mathbf{pa}_i)$. This assumption is standard in identifiability analysis to separate structural identification from functional identification \citep{peters14a-resit-paper}.
\end{assumption}

\subsection{Proof of Theorem \ref{thm:linearization}}
\label{app:proof_theorem1}

We provide the proof for the linearization of endogenous variables under component-wise distortion.

\begin{proof}
Let $s_i$ be the true endogenous variable generated by $s_i = h_i(\mathbf{pa}_i) + n_i$. Let $z_i = \psi_i(s_i)$ be the transformed variable. We assume the learned representation $z$ constitutes a valid ANM satisfying Eq.~\ref{eq:learned_anm}:
\begin{equation*}
    z_i = f_i(\widehat{\mathbf{pa}}_i) + \epsilon_i, \quad \text{where } \epsilon_i \indep \widehat{\mathbf{pa}}_i.
\end{equation*}

\textbf{1. Gradient of the Learned Model:}
By the assumption of Structural Consistency, the learned parents are transformations of the true parents: $\widehat{\mathbf{pa}}_i = \boldsymbol{\psi}_{\mathbf{pa}_i}(\mathbf{pa}_i)$. Since $\epsilon_i$ represents the independent exogenous noise in the learned ANM, it holds no functional dependence on the ancestors $\mathbf{pa}_i$. Therefore, $\nabla_{\mathbf{pa}_i} \epsilon_i = \mathbf{0}$.

We differentiate the learned model expression for $z_i$ with respect to the true parents $\mathbf{pa}_i$:
\begin{equation}
    \nabla_{\mathbf{pa}_i} z_i = \nabla_{\mathbf{pa}_i} [f_i(\widehat{\mathbf{pa}}_i) + \epsilon_i] = \nabla_{\widehat{\mathbf{pa}}_i} f_i(\widehat{\mathbf{pa}}_i) \cdot \nabla_{\mathbf{pa}_i} \boldsymbol{\psi}_{\mathbf{pa}_i}(\mathbf{pa}_i).
\end{equation}



\textbf{2. Gradient of the True Process:}
We evaluate the gradient on the true generative process $z_i = \psi_i(h_i(\mathbf{pa}_i) + n_i)$ using the chain rule:
\begin{equation}
    \nabla_{\mathbf{pa}_i} z_i = \psi_i'(h_i(\mathbf{pa}_i) + n_i) \cdot \nabla_{\mathbf{pa}_i} h_i(\mathbf{pa}_i)
\end{equation}

\textbf{3. Equating Dependencies:}
Equating the two expressions for $\nabla_{\mathbf{pa}_i} z_i$:
\begin{equation}
    \psi_i'(h_i(\mathbf{pa}_i) + n_i) \cdot \nabla_{\mathbf{pa}_i} h_i(\mathbf{pa}_i) = \nabla_{\widehat{\mathbf{pa}}_i} f_i(\widehat{\mathbf{pa}}_i) \cdot \nabla_{\mathbf{pa}_i} \boldsymbol{\psi}_{\mathbf{pa}_i}(\mathbf{pa}_i)
\end{equation}
Since $s_i$ is endogenous, $\nabla_{\mathbf{pa}_i} h_i(\mathbf{pa}_i) \neq \mathbf{0}$ almost everywhere. We project onto the gradient direction to isolate the transformation derivative:
\begin{equation}
    \psi_i'(h_i(\mathbf{pa}_i) + n_i) = \frac{\nabla_{\widehat{\mathbf{pa}}_i} f_i(\widehat{\mathbf{pa}}_i) \cdot \nabla_{\mathbf{pa}_i} \boldsymbol{\psi}_{\mathbf{pa}_i}(\mathbf{pa}_i) \cdot (\nabla_{\mathbf{pa}_i} h_i(\mathbf{pa}_i))^\top}{\|\nabla_{\mathbf{pa}_i} h_i(\mathbf{pa}_i) \|^2}
\end{equation}

\textbf{4. Functional Constraint via Noise Independence:}
We analyze the dependencies of the terms in the projection equation above. Observe that every term on the Right Hand Side (RHS) depends exclusively on $\mathbf{pa}_i$ and is functionally independent of $n_i$. Consequently, the partial derivative of the RHS with respect to $n_i$ is zero. Differentiating the LHS with respect to $n_i$ yields:
\begin{equation}
    \frac{\partial}{\partial n_i} \left[ \psi_i'(h_i(\mathbf{pa}_i) + n_i) \right] = \psi_i''(s_i) \cdot \frac{\partial s_i}{\partial n_i} = \psi_i''(s_i) \cdot 1 = 0
\end{equation}
The condition $\psi_i''(s_i) = 0$ for all $s_i$ implies that $\psi_i$ must have a constant first derivative. Thus, $\psi_i$ is restricted to be an affine transformation $\psi_i(s_i) = a_i s_i + b_i$.
\end{proof}

\subsection{Counter-Example Construction for Proposition~\ref{prop:mixing_impossibility}}
\label{app:proof_prop1}

Consider the specific case where the encoder approximates the causal mechanism $h_2(\cdot)$ and learns the following transformation $\boldsymbol{\Psi}$:
\begin{align}
    z_1 &= s_1 \\
    z_2 &= s_2 - h_2(s_1)
\end{align}
Substituting the structural equation $s_2 = h_2(s_1) + n_2$ into the transformation for $z_2$:
\begin{equation}
    z_2 = (h_2(s_1) + n_2) - h_2(s_1) = n_2
\end{equation}
In this learned representation, $z_1 = n_1$ and $z_2 = n_2$. Since the true noise terms are mutually independent ($n_1 \indep n_2$), the learned variables $z_1$ and $z_2$ are statistically independent. An unsupervised model optimizing for the independence of residuals will accept this solution as valid, identifying a graph with no edges ($z_1 \quad z_2$) rather than the true graph ($z_1 \to z_2$).

\section{Network Architectures}
\label{app:architecture}

To ensure a fair comparison and isolate the contribution of the structural causal modules, we utilize the same backbone encoder and decoder architectures for all models (LANCA, CausalVAE, $\beta$-VAE, and FactorVAE). The input images are resized to $96 \times 96$ pixels with $C=3$ channels. We use Rectified Linear Units (ReLU) as activation functions for all hidden layers in the backbone.

\subsection{Shared Encoder Architecture}
The encoder consists of a 6-layer Convolutional Neural Network (CNN) that compresses the input image into a flat feature vector. The final convolutional layer uses a $3 \times 3$ kernel with no padding to reduce the spatial dimension to $1 \times 1$ before flattening.

\begin{table}[h]
    \centering
    \caption{Architecture of the Shared Encoder. $C$ denotes input channels (3), $K$ denotes kernel size, $S$ denotes stride, and $P$ denotes padding.}
    \label{tab:encoder_arch}
    \begin{tabular}{lcccc}
        \toprule
        \textbf{Layer Type} & \textbf{Parameters} & \textbf{Input Shape} & \textbf{Output Shape} & \textbf{Activation} \\
        \midrule
        Input & - & $96 \times 96 \times C$ & - & - \\
        Conv2d & $K=4, S=2, P=1$ & $96 \times 96 \times C$ & $48 \times 48 \times 32$ & ReLU \\
        Conv2d & $K=4, S=2, P=1$ & $48 \times 48 \times 32$ & $24 \times 24 \times 32$ & ReLU \\
        Conv2d & $K=4, S=2, P=1$ & $24 \times 24 \times 32$ & $12 \times 12 \times 64$ & ReLU \\
        Conv2d & $K=4, S=2, P=1$ & $12 \times 12 \times 64$ & $6 \times 6 \times 64$ & ReLU \\
        Conv2d & $K=4, S=2, P=1$ & $6 \times 6 \times 64$ & $3 \times 3 \times 128$ & ReLU \\
        Conv2d & $K=3, S=1, P=0$ & $3 \times 3 \times 128$ & $1 \times 1 \times 128$ & ReLU \\
        Flatten & - & $1 \times 1 \times 128$ & 128 & - \\
        Linear & - & 128 & $z_{\text{dim}}$ & Linear \\
        \bottomrule
    \end{tabular}
\end{table}

\subsection{Shared Decoder Architecture}
The decoder mirrors the encoder, utilizing Transposed Convolutions to upsample the latent vector back to the original image resolution. The output of the final layer represents the raw logits, to which a Sigmoid function is applied during loss computation (BCE or MSE).

\begin{table}[h]
    \centering
    \caption{Architecture of the Shared Decoder.}
    \label{tab:decoder_arch}
    \begin{tabular}{lcccc}
        \toprule
        \textbf{Layer Type} & \textbf{Parameters} & \textbf{Input Shape} & \textbf{Output Shape} & \textbf{Activation} \\
        \midrule
        Linear & - & $z_{\text{dim}}$ & 128 & - \\
        Reshape & - & 128 & $1 \times 1 \times 128$ & - \\
        ConvTranspose2d & $K=3, S=1, P=0$ & $1 \times 1 \times 128$ & $3 \times 3 \times 128$ & ReLU \\
        ConvTranspose2d & $K=4, S=2, P=1$ & $3 \times 3 \times 128$ & $6 \times 6 \times 64$ & ReLU \\
        ConvTranspose2d & $K=4, S=2, P=1$ & $6 \times 6 \times 64$ & $12 \times 12 \times 64$ & ReLU \\
        ConvTranspose2d & $K=4, S=2, P=1$ & $12 \times 12 \times 64$ & $24 \times 24 \times 32$ & ReLU \\
        ConvTranspose2d & $K=4, S=2, P=1$ & $24 \times 24 \times 32$ & $48 \times 48 \times 32$ & ReLU \\
        ConvTranspose2d & $K=4, S=2, P=1$ & $48 \times 48 \times 32$ & $96 \times 96 \times C$ & Linear \\
        \bottomrule
    \end{tabular}
\end{table}

\subsection{Additive Noise Model (ANM) Layer}
For LANCA, the ANM mechanisms $\{f_i\}_{i=1}^n$ are parameterized by Multi-Layer Perceptrons (MLPs). Each causal variable $z_i$ has a dedicated MLP that takes the masked parents as input.

\begin{itemize}
    \item \textbf{Input:} $z \in \mathbb{R}^n$ (masked by adjacency $\mathbf{A}$)
    \item \textbf{Hidden Layer 1:} Linear($n \to 64$), Activation.
    \item \textbf{Hidden Layer 2:} Linear($64 \to 64$), Activation.
    \item \textbf{Output Layer:} Linear($64 \to 1$).
\end{itemize}

The non-linear activation function is treated as a hyperparameter to ensure both expressiveness and differentiability (required for Theorem 1). We search over the set $\{\text{Tanh}, \text{SiLU}, \text{GELU}\}$.

\section{Hyperparameter Tuning and Model Selection}
\label{app:tuning}

\subsection{Optimization Strategy Details}
\label{app:optimization_strategy}

To mitigate the instability of simultaneous structure and mechanism learning, we employ a three-pronged optimization schedule:

\begin{enumerate}
    \item \textbf{Differential Learning Rates:} We disentangle the learning dynamics by assigning distinct learning rates to the functional parameters ($\eta_{\text{main}}$), the edge weights $\mathbf{U}$ ($\eta_{\text{edge}}$), and the permutation scores $\mathbf{\Pi}$ ($\eta_{\text{perm}}$). Empirically, setting $\eta_{\text{perm}} > \eta_{\text{edge}}$ encourages the model to resolve the topological ordering before fine-tuning specific edge dependencies.
    
    \item \textbf{Temperature Annealing:} We anneal the edge temperature $\tau_{\text{edges}}$ from a high value (soft, high-variance gradients) to a low value (hard, nearly discrete). This allows the model to explore the space of DAGs globally in early epochs before settling into a specific topology.
    
    \item \textbf{Sparsity Warmup:} The regularization weight $\gamma_1$ (for $\mathcal{L}_{\text{sparse}}$) is held at zero during the initial phase ("warmup epochs"). This prevents the strong sparsity prior from suppressing all edges before the SCM has learned meaningful causal mechanisms.
\end{enumerate}

\subsection{Implementation Details}
\label{app:implementation_details}

\paragraph{Latent Dimensions.}
For the synthetic datasets (\textit{Pendulum} and \textit{Flow}), we set the latent dimension $z_\text{dim}=4$, matching the known number of causal variables. For \textit{CANDLE}, we set $z_\text{dim}=12$, despite the dataset having only 6 ground-truth concepts. This over-parameterization reflects realistic scenarios where the true number of underlying factors is unknown. Furthermore, given that the factors in \textit{CANDLE} are not mutually independent but related via confounders, the additional capacity allows the model to potentially capture both the target concepts and the confounding mechanisms.

\paragraph{Model Selection Strategy.}
We acknowledge that unsupervised model selection remains a significant open problem \citep{locatello19a-assumptions}. To ensure a rigorous comparison of model capabilities, we employ a systematic ``oracle validation'' strategy. We sweep hyperparameters and select the best configurations based on ground-truth metrics:
\begin{itemize}
    \item \textbf{Independence Baselines:} Selected based on Mutual Information Gap (MIG).
    \item \textbf{Causal Models (Physics):} Selected based on Mean Mutual Information (MMI) of matched pairs.
    \item \textbf{CANDLE:} Selected based on Unconfoundedness (UC, $\rho=5$).
\end{itemize}

\subsection{Oracle Validation Strategy}
We acknowledge that unsupervised model selection remains a significant open problem. As demonstrated by \citet{locatello19a-assumptions}, consistent model selection is theoretically impossible without access to ground-truth labels or specific inductive biases. To ensure a rigorous and fair comparison of model \textit{capabilities} rather than selection heuristics, we employ a systematic ``oracle validation'' strategy.

Specifically, we utilize ground-truth metrics solely for hyperparameter tuning. This approach isolates the architectural performance from the model selection dilemma. The selection metric varies by model class and dataset to reflect the specific objective of that architecture:

\begin{itemize}
    \item \textbf{Independence Baselines ($\beta$-VAE, FactorVAE):} We select models based on the Mutual Information Gap (MIG), as these models explicitly optimize for statistical independence.
    \item \textbf{Causal Models (LANCA, CausalVAE):} On synthetic datasets with known graph topology (\textit{Pendulum}, \textit{Flow}), we select models based on the Mean Mutual Information (MMI) between matched latent-factor pairs. This ensures we select models that capture the true factors, regardless of the graph structure learned.
    \item \textbf{CANDLE (Real-world):} Due to the lack of independent ground-truth factors, we select based on the Counterfactual Generativeness score (UC, $\rho=5$), which measures the model's ability to perform valid interventions.
\end{itemize}

\subsection{Automated Selection Funnel}
To remove human bias from the tuning process, we implemented an automated three-stage selection funnel. This pipeline progressively filters configurations based on performance and stability.

\paragraph{Stage 1: Exploration.}
We perform a broad sweep over key hyperparameters. To manage computational resources, we utilize Grid Search for spaces with $<50$ combinations and Random Search for larger spaces.
\begin{itemize}
    \item \textbf{Budget:} 100 Epochs.
    \item \textbf{Seeds:} 1 random seed per configuration.
    \item \textbf{Objective:} Identify promising regions in the hyperparameter space.
\end{itemize}

\paragraph{Stage 2: Verification.}
To filter out unstable runs (a common issue in adversarial or structural learning), we select the top-3 configurations from Stage 1 based on the oracle metric.
\begin{itemize}
    \item \textbf{Budget:} 200 Epochs.
    \item \textbf{Seeds:} 3 random seeds per configuration.
    \item \textbf{Selection:} The configurations are ranked by their mean performance across the 3 seeds.
\end{itemize}

\paragraph{Stage 3: Confirmation.}
The single best configuration identified in Stage 2 is subjected to a final, rigorous evaluation to generate the results reported in the main paper.
\begin{itemize}
    \item \textbf{Budget:} 1000 Epochs (ensuring convergence). Most models don't reach 1000 due to the early stopping policy.
    \item \textbf{Seeds:} 10 random seeds.
    \item \textbf{Output:} The mean and standard deviation of all metrics reported in the Result tables.
\end{itemize}

\subsection{Hyperparameter Sweeps}
\label{app:hyperparameters}
Tables \ref{tab:hyperparams_synthetic} and \ref{tab:hyperparams_candle} detail the specific ranges explored for LANCA and the baselines during Stage 1. We adjusted the architecture capacity and regularization strengths for the CANDLE dataset to account for its higher visual complexity and the presence of bounding box supervision.

\begin{table}[!ht]
    \centering
    \caption{\textbf{Synthetic Environments (Pendulum \& Flow).} Hyperparameter search space used for Stage 1 exploration.}
    \label{tab:hyperparams_synthetic}
    \resizebox{0.75\linewidth}{!}{
    \begin{tabular}{llc}
        \toprule
        \textbf{Model} & \textbf{Hyperparameter} & \textbf{Search Space} \\
        \midrule
        \multirow{2}{*}{Common} & Learning Rate & $\{0.01, 0.001, 0.0005\}$ \\
        & Batch Size & $64$ \\
        \midrule
        $\beta$-VAE & $\beta$ & $\{1.0, 2.0, 4.0, 8.0, 16.0, 32.0\}$ \\
        \midrule
        FactorVAE & $\gamma$ (TC Weight) & $\{10.0, 20.0, 40.0, 80.0\}$ \\
        \midrule
        \multirow{3}{*}{CausalVAE} & $\gamma$ (DAG Penalty) & $\{1.0, 5.0, 10.0, 50.0\}$ \\
        & $\beta$ (KL Weight) & $\{1.0, 4.0\}$ \\
        & LR DAG & $\{0.01, 0.001\}$ \\
        \midrule
        \multirow{12}{*}{LANCA (Ours)} & $\lambda_{\text{DAG}}$ (Sparsity) & $\{0.01, 0.05, 0.1\}$ \\
        & $\lambda_{\text{prior}}$ (Independence) & $\{5.0, 10.0, 50.0\}$ \\
        & $\lambda_{\text{recon\_scm}}$ & $\{0.1, 1.0, 5.0, 10.0\}$ \\
        & $\lambda_{\text{sort}}$ (Permutation) & $\{0.1, 1.0, 5.0, 10.0\}$ \\
        & SCM Activation & $\{\text{Tanh}, \text{SiLU}, \text{GELU}\}$ \\
        & SCM Hidden Dim & $\{8, 16\}$ \\
        & Latent Dim ($z_{\text{dim}}$) & $\{32, 64\}$ \\
        & $\tau_{\text{edges}}$ (Start Temp.) & $\{2.0, 5.0, 10.0\}$ \\
        & Warmup Epochs & $\{10, 30, 50\}$ \\
        & Sparsity Delay & $\{0.2, 0.5, 0.8\}$ \\
        & LR DAG Permutation & $\{0.005, 0.01\}$ \\
        & LR DAG Edges & $\{0.005, 0.001\}$ \\
        \bottomrule
    \end{tabular}
    }
\end{table}

\begin{table}[!ht]
    \centering
    \caption{\textbf{Real-World Environment (CANDLE).} Hyperparameter search space used for Stage 1. Note the increased model capacity and inclusion of bounding box regularization ($\lambda_{\text{bb}}$).}
    \label{tab:hyperparams_candle}
    \resizebox{0.75\linewidth}{!}{
    \begin{tabular}{llc}
        \toprule
        \textbf{Model} & \textbf{Hyperparameter} & \textbf{Search Space} \\
        \midrule
        \multirow{2}{*}{Common} & Learning Rate & $\{0.01, 0.001, 0.0005\}$ \\
        & $\lambda_{\text{bb}}$ (BBox Loss) & $\{1.0, 5.0, 10.0\}$ \\
        & Batch Size & $128$ \\
        \midrule
        $\beta$-VAE & $\beta$ & $\{1.0, 2.0, 4.0, 8.0, 16.0, 32.0\}$ \\
        \midrule
        FactorVAE & $\gamma$ (TC Weight) & $\{10.0, 20.0, 40.0, 80.0\}$ \\
        \midrule
        \multirow{2}{*}{CausalVAE} & $\gamma$ (DAG Penalty) & $\{1.0, 5.0, 10.0, 50.0\}$ \\
        & $\beta$ (KL Weight) & $\{1.0, 4.0\}$ \\
        \midrule
        \multirow{10}{*}{LANCA (Ours)} & $\lambda_{\text{DAG}}$ (Sparsity) & $\{0.01, 0.05, 0.1\}$ \\
        & $\lambda_{\text{prior}}$ (Independence) & $\{5.0, 10.0, 50.0\}$ \\
        & $\lambda_{\text{recon\_scm}}$ & $\{0.1, 1.0, 5.0, 10.0\}$ \\
        & $\lambda_{\text{sort}}$ (Permutation) & $\{1.0, 5.0\}$ \\
        & SCM Activation & $\{\text{Tanh}, \text{SiLU}, \text{GELU}\}$ \\
        & SCM Hidden Dim & $\{16, 32\}$ \\
        & Latent Dim ($z_{\text{dim}}$) & $\{64, 128\}$ \\
        & $\tau_{\text{edges}}$ (Start Temp.) & $\{2.0, 5.0, 10.0\}$ \\
        & Warmup Epochs & $\{20, 40\}$ \\
        & Sparsity Delay & $\{0.0, 0.5, 0.8\}$ \\
        \bottomrule
    \end{tabular}
    }
\end{table}

\end{document}